\documentclass[sigconf,nonacm,anonymous=false]{acmart}
\newenvironment{Tabular}[2][1]
{\def\arraystretch{#1}\tabular{#2}}
{\endtabular}
\usepackage{graphicx}
\usepackage{caption}
\usepackage{makecell}
\usepackage{amsmath}
\DeclareMathOperator*{\argmin}{arg\,min}

\newcommand{\norm}[1]{\left\lVert#1\right\rVert}

\usepackage{gensymb}

\usepackage{subfig}
\usepackage{algpseudocode}
\usepackage{algorithm}
\usepackage{cleveref}
\usepackage{siunitx}

 \usepackage{bm} 
 \usepackage{tikz}
\usepackage{hhline}

\usepackage{placeins}

\usepackage{tabularx,booktabs}
\newcolumntype{Y}{>{\centering\arraybackslash}X}

\usepackage{pgfplots, pgfplotstable,booktabs}
\usetikzlibrary{positioning}

\AtBeginDocument{%
  \providecommand\BibTeX{{%
    \normalfont B\kern-0.5em{\scshape i\kern-0.25em b}\kern-0.8em\TeX}}}

\setcopyright{acmcopyright}
\copyrightyear{2023}
\acmYear{2023}
\acmDOI{XXXXXXX.XXXXXXX}

\acmConference[arXiv]{Preprint}{November 30}{2023}
\acmPrice{15.00}
\acmISBN{978-1-4503-XXXX-X/18/06}

\begin{document}

\title{Physical Adversarial Examples for Multi-Camera Systems}

\author{Ana R\u{a}du\c{t}oiu}
\orcid{0000-0002-3139-2954}
\affiliation{%
  \institution{Technical University of Munich}
  \streetaddress{Arcisstraße 21}
  \city{Munich}
  \country{Germany}
  \postcode{80333}
}
\email{anaradutoiu@gmail.com}

\author{Jan-Philipp Schulze}
\orcid{0000-0003-1787-4102}
\affiliation{%
  \institution{Technical University of Munich}
  \streetaddress{Arcisstraße 21}
  \city{Munich}
  \country{Germany}}
\email{jan-philipp.schulze@tum.de}

\author{Philip Sperl}
\orcid{0000-0002-7901-7168}
\affiliation{%
  \institution{Technical University of Munich}
  \city{Munich}
  \country{Germany}
}
\affiliation{
    \institution{Fraunhofer AISEC}
    \city{Garching}
    \country{Germany}
}
\email{philip.sperl@aisec.fraunhofer.de}

\author{Konstantin B\"{o}ttinger}
\orcid{0000-0002-9337-7506}
\affiliation{%
 \institution{Fraunhofer AISEC}
 \city{Garching}
 \country{Germany}}
\email{konstantin.boettinger@aisec.fraunhofer.de}

\renewcommand{\shortauthors}{R\u{a}du\c{t}oiu et al.}

\begin{abstract}
  Neural networks build the foundation of several intelligent systems, which, however, are known to be easily fooled by adversarial examples.
  Recent advances made these attacks possible even in air-gapped scenarios, where the autonomous system observes its surroundings by, e.g., a camera.
  We extend these ideas in our research and evaluate the robustness of multi-camera setups against such physical adversarial examples.
  This scenario becomes ever more important with the rise in popularity of autonomous vehicles, which fuse the information of several cameras for their driving decision.
  While we find that multi-camera setups provide some robustness towards past attack methods, we see that this advantage reduces when optimizing on multiple perspectives at once.
  We propose a novel attack method that we call Transcender-MC, where we incorporate online 3D renderings and perspective projections in the training process.
  Moreover, we motivate that certain data augmentation techniques can facilitate the generation of successful adversarial examples even further.
  Transcender-MC is 11\% more effective in successfully attacking multi-camera setups than state-of-the-art methods.
  Our findings offer valuable insights regarding the resilience of object detection in a setup with multiple cameras and motivate the need of developing adequate defense mechanisms against them. 
\end{abstract}




\maketitle

\section{Introduction}
Ground-breaking Machine Learning (ML) algorithms show unprecedented performance in different application areas, such as computer vision \cite{diffusion_cv}, medical software \cite{medical} or language modeling \cite{language_models}, and often security and privacy concerns arise as an aftermath.
For instance, adversarial examples, which were first introduced by Szegedy et al. \cite{szegedy2014intriguing}, can be described as manipulated inputs that fool models to deliver an incorrect prediction. 
State-of-the-art (SOTA) techniques allow us to create adversarial examples against various Neural Network (NN) models \cite{biggio2023security}. 
This represents a significant risk, as many intelligent systems employ Deep Learning (DL) models as an integral part. 

A significant threat is represented by physical adversarial examples (PAEs), which are modified inputs that are physically materialized, and can therefore be used to influence the output of NNs that process information from the real world \cite{wei2022physically}. 
For example, object detectors \cite{yolov3, fast-rcnn}, which are a foundational element of many vision systems, can be fooled by an adversarial patch that is crafted to be identified as an object belonging to a class chosen by the attacker. 
This is especially concerning in scenarios like autonomous driving, where the information processed by object detectors is used in the decision-making process of the car.

PAEs have to deal with challenges that may appear in the real world, such as various camera angles, changing lighting conditions and occlusions. 
Chen et al. \cite{shapeshifter} were the first to propose a method for generating PAEs against the monocular object detector Faster R-CNN \cite{fast-rcnn}. Their method, called ShapeShifter, is based on digitally modeling the perturbations that can appear in a physical environment.
Following them, many other research works developed adversarial examples against single-view detectors \cite{wang2022survey}. 
However, numerous intelligent vision platforms use an ensemble of cameras to gather video input from their surroundings, such as autonomous cars that use a multi-camera setup to acquire a $360^{\circ}$ field of view. 

There is little research on the impact of adversarial patches against a vision system that uses multiple cameras, implying that we have an incomplete understanding of the potential vulnerabilities of such platforms. 
When used in such a setup, a PAE has to be robust from multiple angles at the same time. 
This adds a layer of complexity, as it represents an additional requirement that adversarial patches need to satisfy.

In this work, we investigate the robustness of 2D adversarial patches against a multi-camera object detection system.
A 2D adversarial example is a resource-efficient form of attack, which is easy to optimize and materialize, compared to 3D camouflages or 3D printed objects. 
Hence, it might be a preferred approach for many attackers, motivating the research interest of our work. 
We propose an extension of ShapeShifter, called Transcender-MultiCamera (Transcender-MC), that optimizes the 2D attack to be robust against a multi-camera setup.

Our contributions are the following: 
\begin{itemize}
    \item We propose a new method for generating adversarial patches, called Transcender-MC, which uses 3D renderings and out-of-plane camera motions to improve the robustness of 2D attacks in real-life environments. 
    Transcender-MC also introduces a multi-camera specific training data augmentation technique for optimizing patches that are robust from multiple angles, at the same time.  
    \item We present a comparative evaluation of Transcender-MC and ShapeShifter. 
    \item We assess the optimized patches in a lab environment, where we employ an ensemble of three cameras, inspired by the vision system of a self-driving car. 
    We introduce new metrics for evaluating the effectiveness of PAEs in multi-camera setups and propose 3D rendering as approach for evaluating PAEs. 
    \item We show that a multi-camera setup does not inherently offer protection against PAE. 
\end{itemize}

\section{Related Work}

\subsection{Object Detection}
The backbone of intelligent vision systems is represented by object detectors, which have the task of identifying and classifying objects in an image.
The detector's output is comprised of a bounding box, which encloses the object, and a classification score for the identified object. 
Initially, monocular object detectors that output 2D bounding boxes, such as YOLOv3 \cite{yolov3} and Fast-RCNN \cite{fast-rcnn}, were developed. 
Afterwards, 3D object detectors were proposed, which add a third dimension, the depth, to the bounding box.
Since monocular camera-based 3D detectors have a high error in estimating the depth, multi-view alternatives were investigated. 
Steoreo-based 3D detectors leverage the 2D detections on the two images to infer the depth information \cite{li2019stereo}.
Based on the stereo matching technique, other models create a Pseudo-LiDAR representation of the data and perform the 3D detection on a point cloud. 
Since self-driving vehicles use multiple cameras to acquire environment information, novel methods employ multi-view 3D detection. 
For example, Lisft-Splat-Shoot (LSS) \cite{lift-splat-shoot} is a technique of representing the surroundings in a Bird's Eye View (BEV) representation. 
The backbone on the LSS method is comprised of extracting 2D features of each image of the multi-view dataset. 
Then, the features are aggregated in a 3D frustum, which is then splatted into a BEV surface.


\subsection{Adversarial Examples in the 3D Environment}
Adversarial examples are manipulated inputs that aim at changing the output of an inference model. 
Physical adversarial examples that target object detectors have to be robust enough to fool a detector under various circumstances. 
Previous work approach the topic of adversarial patches from multiple views. 
While some methods focus on generating adversarial camouflages \cite{duan2020adversarial,den2020adversarial, FCA, CAC, camou}, in this work we center on adversarial 2D patches only. 

Although the PAE is two-dimensional, in a real-life scenario we encounter various viewing angles and distortions.
Several generation methods for physical adversarial patches consider the implications of using a planar adversarial example in a three-dimensional world. 
Lennon et al. \cite{Lennon_2021_ICCV} study the effectiveness of adversarial patches from different out-of-plane camera positions against a 2D object detector. 
The authors offer a qualitative assessment regarding how the robustness can be increased using certain 3D camera movements in the training routine. This work also showcases the influence of the target class of an attack on its robustness. 
Tarchoun et al. \cite{tarchoun_1} also study how the camera angles can influence the robustness of the adversarial patch.
They train the adversarial patch using a single view of a multi-view dataset and then project the optimized attack on the rest of the corresponding views. 
They discover that variations in the viewing angle have a detrimental effect on the patch effectiveness. 
In a different work \cite{tarchoun_2}, the same authors study the effectiveness of a digital adversarial patch against a multi-view object detector. 
They conclude that an adversarial patch optimized to attack a single-view detector performs poorly in a multi-view setting and report that the effectiveness of the attack decreases as the number of attacked views from a multi-view dataset is decreased. 
Their evaluation supports the idea that, by using multi-view detectors, the effectiveness of PAE in a multi-camera setup is decreased. 
Hoory et al. \cite{hoory2020dynamic} propose a method for creating adversarial patches that are robust from different angles by designing an algorithm that optimizes dynamic patches, that change according to the position of the camera. 
They generate a changing PAE that is displayed on screens attached to a car, advocating that their method has a better efficiency for a multi-view scenario compared to using static patches. 
Yael et al. \cite{mathov2022enhancing} try to solve the problem of varying camera angles and deformations by constructing a 3D replica of the evaluation environment. 
They render images of a scene and use them at training time by projecting the adversarial patch on a support object. At test time, the PAE is physically printed on the object and a moving camera is connected to a 2D object detector. 
Based on their evaluations, they conclude that the method is able to produce robust attacks from multiple views.  

Other works investigate the benefit of adding 3D elements in the training pipeline to leverage attacks against image classifiers. Byun et al. \cite{byun2022transferability} improve the transferability of adversarial examples against a classifier by applying the perturbed images on different objects and capturing the scene from different perspectives during training. In their work, the authors also showcase that this method is efficient for increasing the robustness of attacks against face verification models.

Table \ref{tab:related_work_limit} presents the contributions of related work on adversarial examples and limitations which we address in this paper. 

\begin{table}[h!]
\centering
\begin{tabular}{l c c c c c c c }
\Xhline{2\arrayrulewidth}
     \textbf{Contribution} & \cite{Lennon_2021_ICCV} & \cite{tarchoun_1} & \cite{tarchoun_2} & \cite{hoory2020dynamic} & \cite{mathov2022enhancing} & \cite{byun2022transferability} & Ours \\
     \hline 
     Physical Attack & - & - & - & \checkmark & \checkmark  & - & \checkmark \\ 
     Multi-Cam. Set. & - & \checkmark & \checkmark & - & -  & - & \checkmark\\
     Use 3D Obj. & - & - & - & - & \checkmark & \checkmark & \checkmark\\ 
     Diff. Rendering & - & - & - & - & -  & - & \checkmark \\ 
     3D Cam. Angles & \checkmark & \checkmark & \checkmark & \checkmark & \checkmark & \checkmark & \checkmark \\
\Xhline{2\arrayrulewidth}
\end{tabular}
\caption{Contributions and limitations of related work.}
\label{tab:related_work_limit}
\end{table}

\section{Preliminaries}

\subsection{Threat Model} \label{subsection:threat_model}
Self-driving cars are gaining popularity and becoming increasingly prevalent in traffic. 
Inspecting how vulnerable their vision systems are against attacks is a step towards enforcing their trustworthiness and ensuring the safety of road participants. 

\begin{figure}[t]
    \centering
    \includegraphics[scale=0.8]{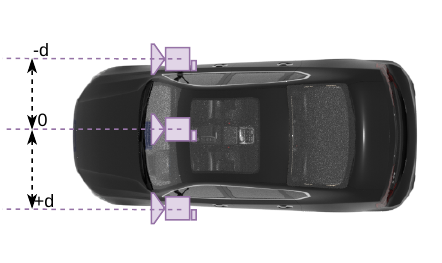}
    \caption{
    Our assumed multi-camera setup, inspired by the vision system of an autonomous car. Ref? 
    }
    \label{fig:multicamera-setup}
\end{figure}

Considering this, we assume an attacker that tries to create a PAE against the object detection system of a self-driving car. 
Inspired by intelligent vision platforms like Tesla Autopilot \footnote{https://www.tesla.com/autopilot}, we assume a setup with three video devices that emulate the frontal cameras of a vehicle. 
The three cameras are positioned at the same height relative to the ground and on a straight line that connects the side mirrors, as represented in Fig. \ref{fig:multicamera-setup}. 
Thus, we have a left, central, and right camera, with an offset $d$ between them. 
They are running simultaneously and each forwards the collected video input to a 2D object detector. 
The scope of the attacker is to trick all three cameras at the same time, assuming that the attack is placed in their common field of view. 

Besides the resemblance with the multi-camera vision system of autonomous cars, several other reasons also motivate our choice for the presented camera setup.
First, many multi-camera object detectors are either based on 2D detectors, or use as backbone a 2D feature extractor and region proposal networks \cite{qian20223d}. 
Hence, it is important to study how robust the 2D object detectors running in parallel are as an ensemble.
Second, our research should motivate future studies that investigate different setups, to reveal which one offers higher security and find potential correlations.

The attacker wishes to perform a targeted physical adversarial patch and has complete knowledge about the object detector model, pursuing a white-box approach. 
The attacker has limited resources: 
she can only print the patch on a 2D surface, such as a paper, but she cannot create adversarial camouflages or use a 3D printer. 
Therefore, we assume that the attack is printed on a white background. 
Moreover, the printed adversarial patch is fixed in space and, once printed, it cannot be modified.

\subsection{The YOLOv3 Object Detector}
An object detection model has mainly two tasks: 
to regress the coordinates that indicate where objects are present in the image, and to assign each object a class label. 
Some object detectors, like Faster R-CNN \cite{fast-rcnn} accomplish the tasks using two networks, one for region proposal and one for classification inside the proposed regions, and are named accordingly two-stage detectors. 
The other class of detectors, called one-stage detectors, predict the bounding box coordinates and the classification label using one network. YOLOv3 belongs to the second class, being a monocular 2D one-stage detector introduced by Farhadi et al. \cite{yolov3} which has been widely used because of its high inference speed. 
Formally, we can consider YOLOv3 as a function $f(\bm x) = \bm Y^{'}$ which takes as input an image $\bm x$ and outputs a vector $\bm Y^{'}=(\bm b^{'}_{box}, b^{'}_{obj}, b^{'}_{l})$, where $\bm b^{'}_{box} = (b^{'}_{x}, b^{'}_{y}, b^{'}_{h}, b^{'}_{w})$ are the regressed bounding box coordinates ($b^{'}_{x}, b^{'}_{y}$ represent the center coordinates of the box, while $b^{'}_{h}, b^{'}_{w}$ are the width and height of it), $b^{'}_{obj}$ is the objectness score that depicts the probability of an object in the bounding box, and $b^{'}_{l}$ is the inferred class label. In this work, we will also refer to the objectess score as ``detection score''. 
With $\bm Y = (\bm b_{box}, 1, b_{l})$ representing the ground truth information as a vector containing the real bounding box coordinates of an object $\bm b_{box}$, a maximum objectness score of 1 and the real label of the object $b_{l}$, then the YOLOv3 loss function can be described as $\mathcal{L}(f(\bm x), \bm Y)$. 
It fundamentally fulfills two tasks: 
it computes the classification loss for the real label and the output label of the function, as well as the distance between the regressed bounding box and the ground truth box that encompasses the object. 

In this paper, we generate attacks against the YOLOv3 object detector. At test time, the attack is shown to an ensemble of YOLOv3 detectors that receive video input from multiple cameras running in parallel in different positions. 
Each YOLOv3 instance is trained on the COCO2017 dataset \cite{lin2015microsoft}, which contains labeled images of objects from 80 different classes.

\subsection{Expectation over Transformation}
An adversarial example materialized in the physical world needs to be robust enough to overcome the added layers of complexity when transitioning from the digital realm to a real-life environment. 
Different challenges appear in a physical scenario: various camera positions and camera properties, different lighting conditions, occlusions and deformations of the patch. 
An algorithm designed to generate robust PAE that are effective over a distribution of transformation that can appear in a physical environment is the Expectation over Transformation (EoT) method, developed by Athalye et al. \cite{EoT}. 
To tackle the above mentioned challenges, the authors model the variations in positions and in color space using a distribution of transformations $t \in T$ that are applied to the planar adversarial example. 
The final adversarial noise is then obtained by solving the following optimization objective
        

\begin{equation}
\begin{aligned}
    \text{minimize} \quad & \mathop{\mathbb{E}}_{t \sim T}\left[ \norm{t(\bm x) - t(\bm x + \bm \delta)}_{p} \right] \\ 
    \text{subject to} \quad & \mathop{\mathbb{E}}_{t \sim T} \left[ f(t(\bm x + \bm \delta)) \right] = l 
\end{aligned}
\end{equation}
where $\bm x$ is the original image, $\bm \delta$ is the added adversarial perturbation, and $l$ is the label of the targeted attack. 

There are several options for choosing the transformation distribution $T$. 
Ideally, the chained application of transformations $t$ on the attack should mimic how the patch could look like in various scenarios in the real world. 
The work of Sava et al. \cite{sava2022assessing} offers an overview of possible choices for $T$ in related research. 
According to the authors, most methods employ image filters, such as brightness and contrast change, and affine transformations, like shear, rotation and translation. 
Moreover, Sava et al. \cite{sava2022assessing} also discover that the combination of transformations used to optimize a patch influences its robustness, however there is no algorithm for determining the optimal combination for a specific attack scenario.

\subsection{Differentiable Rendering} 
Classical 3D rendering refers to the process of creating 2D images from a 3D representation of a scene. 
Since Kato et al. \cite{kato2018neural} proposed an algorithm that allowed differentiating the rasterization step, 3D rendering was included as an integral part of end-to-end NNs. 
Several research studies on adversarial learning explored the potential of including 3D elements to leverage the robustness of attacks \cite{FCA, byun2022transferability, CAC, maesumi2021learning}. 
While most of the time differentiable rendering in adversarial learning is used to create 3D adversarial coatings, Byun et al. \cite{byun2022transferability} use this technique to improve the transferability of adversarial patches against image classifiers, showing that this technique can also be beneficial in the context of 2D adversarial images.

\section{Methodology}

\begin{figure*}
    \centering
    \includegraphics[scale=0.65]{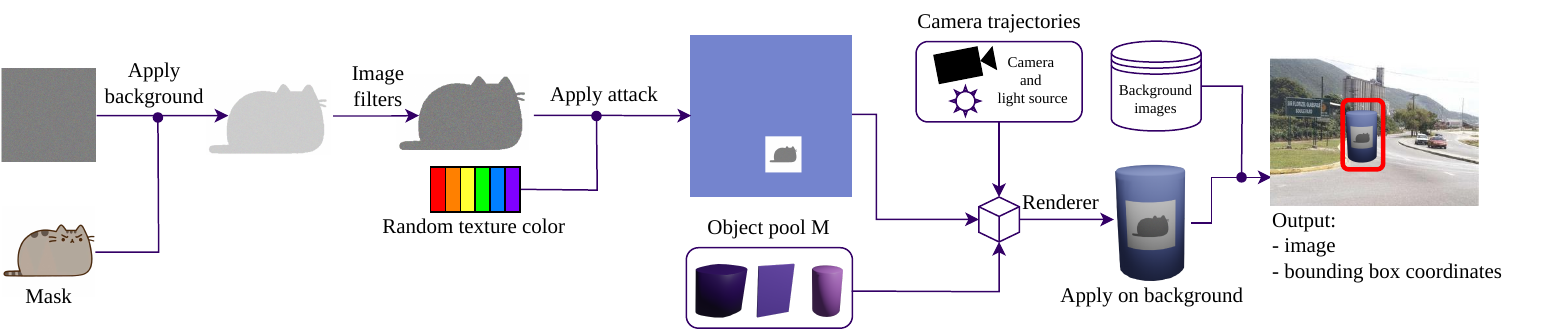}
    \caption{One iteration of applying transformations on the adversarial example.}
    \label{fig:transcender-pipeline}
\end{figure*}

Most previous methods that employ the EoT algorithm for generating PAEs include only image filters and affine transformations in the transformation set $T$. 
However, this poses several disadvantages. 
First, cameras in a physical environment can have various 3D positions and only perspective projections can model how an adversarial patch is perceived from various angles. Moreover, in a physical environment, the adversarial patch will be applied on a 3D object, which has various implications: (1) while the attack is two-dimensional, it becomes part of a three dimensional item and (2) if the 3D object does not have a completely flat surface, the attack will also be deformed.
At the same time, most of the PAE generation methods focus on fooling monocular 2D object detectors and disregard the possibility of a multi-camera setup. 


Thus, we introduce Transcender-MC, an adversarial patch generation algorithm which leverages a 2D patch and transcends it to the third dimension, making it robust in a 3D setting. 
Our proposed method also includes an augmentation of training for multi-camera scenarios that makes the attack effective from multiple angles at the same time. 

\subsection{Patch Generation}

Transcender-MC is a PAE generation method that builds on the training pipeline of ShapeShifter \cite{shapeshifter}.
To accomplish our goal of generating 2D adversarial patches, we are inspired by the idea of Byun et al. \cite{byun2022transferability} of applying the adversarial patch during training on different objects, while also moving the camera in random 3D positions. 

The training pipeline of Transcender-MC is the following: first, random transformations from a distribution of image filters are applied. 
Next, a 3D object is chosen from a pool of meshes and the adversarial example is placed on the texture of the object. 
The virtual camera is then pointed towards the 3D mesh while being randomly rotated and translated. 
Then, the object is rendered and the bounding box coordinates are generated. 
The output of the detector and the target label together with the bounding box coordinates are used to compute the YOLOv3 loss used to update the adversarial example via propagation. 
Since YOLOv3 is a one-stage detector, the entire optimization pipeline is end-to-end differentiable.

Based on this, we can derive a formal description of our optimization objective. 
Let $p = p_{1} \circ p_{2} \circ \hdots \circ p_{n}$ with $p_{k} \in P$ a composition of image filters (brightness, contrast, hue, Gaussian blur) applied on an image $p(\bm{x})$. Let $f_{obj}(\bm{x}) = (\mathbf{b}_{box}^{'}, \ b^{'}_{obj}, \ b^{'}_{cls})$ be an object detector and its output: the four bounding box coordinates, the objectness score and the class label. 
To render the attack on a 3D mesh, we define a renderer $\mathbf{I} = \mathcal{R}(m, \ \mathbf{T}, \ \theta_{cam})$ where $m \in M$ is the mesh from a mesh pool $M$, $\mathbf{T}$ is a texture, $\theta_{cam}$ are the camera parameters $\theta_{c} \in \Theta$ and  $\mathbf{I} \in \mathbb{R}^{h \times w \times 3}$ is the output image. 
We define $\mathbf{T}_{adv} = \mathbf{T} \oplus \bm{x}_{adv}$ the texture after we apply an adversarial example on it and the function $\Phi(\mathbf{I}, \ \mathbf{x})$ that applies the rendered image $\mathbf{I}$  on a background $\bm{x} \in X_{b}$ (we assume that we render the object with adversarial texture on a transparent background). 
Having the ground truth information $\mathbf{Y} = (\bm{b}_{box}, \ 1, \ b_{l})$ where $\bm{b}_{box}$ are the bounding box coordinates, $b_{obj} = 1$ and the target label $b_{l}$, the optimization problem after applying on the initial patch $\bm{x}_{p}$ the Carlini \& Wagner \cite{carlini2017towards} change-of-variable $\bm{w} = \frac{1}{2}(\text{tanh}(\bm{x}_{p}) + 1)$   becomes: 

\begin{multline}
    \underset{\bm{w} \in \mathbb{R}^{h \times w \times 3}} 
         {\argmin} \mathop{\mathbb{E}}_{\begin{subarray}{c}\bm{x} \sim X_{b}\\p \sim P \\ \theta_{c} \sim \Theta \\ m \sim M \end{subarray}}
         \left[ \mathcal{L}(f(\Phi( \mathcal{R}(m,\ \mathbf{T} \oplus p(\bm{w}), \ \phi_{c}) , \ \bm{x}), \mathbf{Y}) \right] \\ 
        + c_{1} \cdot \mathcal{L}_{TV}(\bm{w})
        + c_{2} \cdot \mathcal{L}_{NPS}(\bm{w})
\end{multline}

where we sample the mesh $m \in M$ and camera parameters $\phi_{c} \in \Phi$ from distributions.
$ \mathcal{L}_{TV}$ i the Total Variation (TV) loss which acts as a regularization term that has the scope of denoising the attack and assuring a smooth gradient in pixel colors.
The non-printability score $\ \mathcal{L}_{NPS}$ is the second regularizer, which was first introduced by Sharif et al. \cite{sharif2016accessorize} and it computes the loss between the actual RGB colors of the attack and the set of colors available for a specific printer or display.
In practice, each printer can only represent a subset of all colors, and this loss term ensures that the attack is formed of colors that can be reproduced with fidelity in printed form. 
The distribution of meshes $M$ is represented by objects that the patch can be applied to. 
The camera parameters $\Phi$ define the allowed camera trajectories. 
The texture $\mathbf{T}$ is a random color that the attack is applied onto, in a predefined location, such that, after rendering it, the attack is visible in the final image. 
The optimization problem can be solved using iterative optimization algorithms. 
By employing a differentiable renderer, the entire transformation pipeline is end-to-end differentiable. 
The Transcender-MC pipeline is visually depicted in Fig. \ref{fig:transcender-pipeline}.

\subsection{Training Data Augmentation for Multi-camera Setups} \label{subsection:data_augmentation}
\begin{figure}[t]
    \centering
    \includegraphics[scale=0.4]{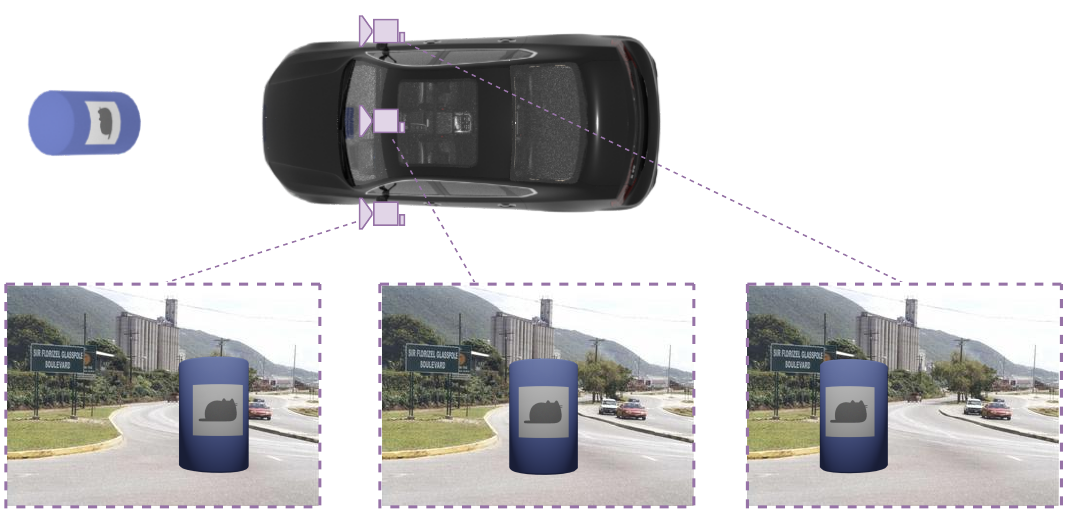}
    \caption{Three renderings of the same 3D object containing an adversarial example, based on setup with three cameras.}
    \label{fig:transcender-mc}
\end{figure}

To adapt to a multi-camera environment, we introduce in our training pipeline images of the attack from different angles. 
This is feasible since we can change the camera parameters and adjust its position to emulate a setup with multiple cameras.
According to our threat model assumptions, the three frontal video devices are placed on a parallel line, which means that each of them has a different perspective of an object that is located in their common field of view. 
This is represented in Fig. \ref{fig:transcender-mc}. 
To emulate this behavior, at training time we render three images of the same scene that contains an adversarial example from three different angles. 
For this, we fix a central camera with an offset $d_{1} = 0$ and sample a random translation parameter $\gamma$. 
The left camera is then placed at $d_{2} = -\gamma$ and the right camera at $d_{3} = \gamma$. 
At each training step, the offset $\gamma$ is sampled from the interval $[0, \delta_{d}]$, such that the adversarial example is always in the common field of view of the three cameras. 
After rendering the three images of the same object, they are placed independently on the same background image. 
We do not to modify the background based on the perspective change, as we believe that the small changes in the environment information have too small of an impact for the optimization of the attack. 
Thus, at the end of one training pipeline we generate three images of the same attack placed on an object and we enforce the optimization to output an attack that is robust in the assumed multi-camera setup. 

\section{Experiments}
In this section, we first introduce our experimental setup and then showcase a comparative evaluation between Transcender-MC and another SOTA method.
Our evaluation in a real-world multi-camera setup indicates the effectiveness of our novel PAE generation algorithm, thus revealing significant security risks for today's autonomous systems. 

In our comparison, we also analyze the performance of our method without using the previously introduced data augmentation technique during training, which we refer to as Transcender.
We further analyze Transcender in our ablation studies at the end of this section.

\subsection{Experimental Setup}

\subsubsection{Evaluation Metrics} 
The design of our evaluation metrics is taking into consideration the previously described camera setup.
Each camera is forwarding its output to one instance of a pretrained YOLOv3 detector. 
In the following, we consider a successful attack to be a patch that is detected in an image as its intended target label with a detection score greater than 50\%. 
We use the following metrics to characterize an adversarial example in a multi-camera setup:
\begin{itemize}
    \item Attack strength:
    describes how many cameras detected the attack. 
    We call a strong attack an instance detected by all cameras, a weak attack an attack detected by two cameras, and a single attack an attack detected by only one camera. 
    We will refer to an attack that fooled no cameras as a failed attack, while one which fooled at least one camera is denoted as a working attack.
    \item Multi-camera robustness score:
    performance metric that takes into consideration the attack strength and the individual detection score of each camera. 
    The score $S$ for an adversarial patch is computed as
    \begin{align}
        S = n_{\text{valid}} \cdot \sum_{i \in \{1,2,3\}}s_{i}
    \end{align}
    where $n_{\text{valid}} \in \{0, 1, 2, 3\}$ is the 
    number of cameras fooled by the attack and $s_{i}$ is the detection score of the \textit{i}-th camera. 
    According to this formula, we give the highest score to patches that are detected by all three cameras. 
\end{itemize}

\subsubsection{Attack Settings}
Each individual adversarial example is defined by the target class, the transformation configuration and the training parameters. 

\paragraph{Target Classes} 
Since we attack YOLOv3 trained on the COCO2017 dataset, the target classes should be among the 80 classes of the dataset. 
In our experiments, we use the following: ``person'', ``car'', ``traffic light'' and ``stop sign''. 
The objects belonging to the four groups can be encountered in a driving scenario and are thus relevant for our use case. 
Moreover, each of the four classes has a different representation in the dataset: ``person'' and ``car'' are over-represented, each with 66,808 and 12,786 appearances, respectively, while ``traffic light'' and ``stop sign'' are under-represented, with 4,330 and 1,803 appearances in the images.
Having different target classes with various numbers of representation in the dataset allows us to observe the influence of the class on the robustness of patches. 
Instead of generating rectangular patches, we increase the stealthiness of the planar attack by applying a mask that shapes it in different forms, such as the outline of an emoji.

\paragraph{Transformation Parameters}
Each attack is based on a configuration of transformations, which represents the transformation distribution $T$. 
The transformations that we include for Transcender-MC and Transcender are presented in Table \ref{table:all_parameters}, while ShapeShifter transformations are presented in Table \ref{table:shapeshifter_parameters}, in the Appendix. 
All three methods apply the same image filters, and their parameter ranges are chosen such that the attack features are still distinguishable. 
Besides the transformations in color space, the adversarial example also undergoes geometrical transformations. 
In a physical environment, the PAE can exhibit different spatial positions and can be rotated at random angles relative to the cameras. 
ShapeShifter uses affine transformations, while Transcender-MC employs rigid-body motions in form of rotations and translations on the camera to realize this.  
The transformation parameters for ShapeShifter, Transcender-MC and Transcender are picked to represent the same attack scenario. 
For example, the resize function parameters of ShapeShifter and the values for the camera distance interval are chosen such that the attack has the same resolution during training for both methods. 
Moreover, the two Transcenders apply the attack on a 3D mesh. 
The set of 3D meshes includes objects with different shapes: ``billboard'' is planar, while the two others are round. Each transformation configuration contains one, two or all three objects.

\begin{table}[h]
\centering
\begin{Tabular}[1.2]{l|lll}
\Xhline{2\arrayrulewidth}
\textbf{Img. Filter} & Mechanism & Param. & Range \\
\hline \hline
 Brightness & $\alpha X$ & $\alpha$: scal. & $[0.5, 1.35]$ \\
Contrast & $\beta (X - \overline{X}) + \overline{X}$ &$\beta$: scal. & $[0.5, 1.35]$ \\
Motion Blur & $X \ast K(\alpha, \theta)$ & $\alpha$: ker. size & [0, 0.02]\\
  &  & $\theta$: angle  & $[0, 360]$ \\
 Gaussian Noise & $X + \mathcal{N}(0, \sigma)$ &$\sigma$: stddev.  & $[0, 1]$ \\
Hue & $hue(X) + \alpha$ & $\alpha:$ scal. & $[-0.5, 0.5]$ \\

\hline \hline 
\textbf{Camera Trans.} & Mechanism & Param. & Range \\
\hline \hline 
Rot: Azimuth & $R_{\varphi}C$ & $\varphi:$ angle & $[-35^{\circ}, 35^{\circ}]$ \\ 
Rot: Elevation & $R_{\theta}C$ & $\theta:$ angle & $[-20^{\circ}, 20^{\circ}]$ \\ 
Horiz. Trans. & $C + T_{\alpha}$ & $\alpha:$ scal. & $[-\delta_{x}, \delta_{x}]$ \\ 
Vert. Trans. & $C + T_{\alpha}$ & $\alpha$: scal. & $[-\delta_{h}, \delta_{h}]$ \\
Distance & $C + T_{\alpha}$ & $\alpha:$ scal. & $[d_{min}, d_{max}]$ \\
\hline \hline 
\textbf{3D Rendering} & Mechanism & Param. & Obj. Set \\ 
\hline \hline
3D Mesh Set & $\mathcal{R}(M)$ & $M$: mesh & \{Barrel, Sign, \\
&&& Billboard\} \\
\Xhline{2\arrayrulewidth}
\end{Tabular}
\caption{
    Parameter choice for Transcender-MC and Transcender. 
    $X$ is the input adversarial patch, $X_{h}, \ X_{w}$ are the its height and width, $\overline{X}$ is the average pixel value, $hue(X)$ is the hue channel of the image, $C$ represents the camera coordinates, $R_{\text{angle}}$ is a rotation matrix, $\delta_{h}, \delta{w}$ are the maximum translation offsets, $\mathcal{R}(\cdot)$ is a differentiable renderer.}
\label{table:all_parameters}
\end{table}

\definecolor{dark_purple}{HTML}{3B0058}
\definecolor{mid_purple}{HTML}{564795}
\definecolor{light_purple}{HTML}{7585cf}

\newcommand\barchartheight{17.55em}
\newcommand\barchartwidth{1.55em}
\newcommand\barspacing{1.1em}
\newcommand\barchartmaxyval{110}
\newcommand\ticklabelpadding{2.6em}

\begin{figure*}
    \centering
\begin{tikzpicture}
    \begin{axis}[
    every node near coord/.append style={font=\footnotesize, color=dark_purple, xshift=-1.45em},
    ybar stacked,
    height=\barchartheight,
    width=\textwidth,
    ymin=0,
    ymax=\barchartmaxyval,
    y label style={at={(axis description cs:0.03, 0.5)}},
    symbolic x coords={All, Traffic Light, Stop Sign, Car, Person},
    xtick={data},
    x tick label style={yshift=-\ticklabelpadding},
    bar width=\barchartwidth,
    bar shift=-\barchartwidth-\barspacing,
    ylabel=Percentage,
    clip=false,
    legend style={cells={anchor=west}, font=\footnotesize, at={(0.5, 0.95)}, anchor=north},
    ]
    \addplot[fill=dark_purple, nodes near coords] coordinates {
        (All, 22) (Traffic Light, 0) (Stop Sign, 2) (Car, 37) (Person, 48)
    };
    \addlegendentry{Strong Attacks}
    \addplot[fill=mid_purple] coordinates {
        (All, 14) (Traffic Light, 0) (Stop Sign, 6) (Car, 21) (Person, 26)
    };
    \addlegendentry{Weak Attacks}
    \addplot[fill=light_purple] coordinates {
        (All, 15) (Traffic Light, 6) (Stop Sign, 12.5) (Car, 25) (Person, 24)
    };
    \addlegendentry{Single Attacks}
    \node[font=\scriptsize, xshift=-3.5em, yshift=-1em, rotate=45] at (axis cs:All,-3) {ShapeShft};
    \node[font=\scriptsize, xshift=-0.6em, yshift=-1em, rotate=45] at (axis cs:All,0) {Transc};
    \node[font=\scriptsize, xshift=1.6em, yshift=-1em, rotate=45] at (axis cs:All,-3) {Transc-MC};

    \node[font=\scriptsize, xshift=-3.5em, yshift=-1em, rotate=45] at (axis cs:Traffic Light,-3) {ShapeShft};
    \node[font=\scriptsize, xshift=-0.6em, yshift=-1em, rotate=45] at (axis cs:Traffic Light,0) {Transc};
    \node[font=\scriptsize, xshift=1.6em, yshift=-1em, rotate=45] at (axis cs:Traffic Light,-3) {Transc-MC};

    \node[font=\scriptsize, xshift=-3.5em, yshift=-1em, rotate=45] at (axis cs:Stop Sign,-3) {ShapeShft};
    \node[font=\scriptsize, xshift=-0.6em, yshift=-1em, rotate=45] at (axis cs:Stop Sign,0) {Transc};
    \node[font=\scriptsize, xshift=1.6em, yshift=-1em, rotate=45] at (axis cs:Stop Sign,-3) {Transc-MC};

    \node[font=\scriptsize, xshift=-3.5em, yshift=-1em, rotate=45] at (axis cs:Car,-3) {ShapeShft};
    \node[font=\scriptsize, xshift=-0.6em, yshift=-1em, rotate=45] at (axis cs:Car,0) {Transc};
    \node[font=\scriptsize, xshift=1.6em, yshift=-1em, rotate=45] at (axis cs:Car,-3) {Transc-MC};

    \node[font=\scriptsize, xshift=-3.5em, yshift=-1em, rotate=45] at (axis cs:Person,-3) {ShapeShft};
    \node[font=\scriptsize, xshift=-0.6em, yshift=-1em, rotate=45] at (axis cs:Person,0) {Transc};
    \node[font=\scriptsize, xshift=1.6em, yshift=-1em, rotate=45] at (axis cs:Person,-3) {Transc-MC};

    \node[font=\footnotesize, xshift=-2.7em] at (axis cs:All,55) {51};
    \node[font=\footnotesize] at (axis cs:All,71) {67};
    \node[font=\footnotesize, xshift=2.7em] at (axis cs:All,75) {71};

    \node[font=\footnotesize, xshift=-2.7em] at (axis cs:Traffic Light,10) {6};
    \node[font=\footnotesize] at (axis cs:Traffic Light,45) {41};
    \node[font=\footnotesize, xshift=2.7em] at (axis cs:Traffic Light,56) {52};

    \node[font=\footnotesize, xshift=-2.7em] at (axis cs:Stop Sign,25) {20.5};
    \node[font=\footnotesize] at (axis cs:Stop Sign,54) {50};
    \node[font=\footnotesize, xshift=2.7em] at (axis cs:Stop Sign,60) {56};

    \node[font=\footnotesize, xshift=-2.7em] at (axis cs:Car,87) {83};
    \node[font=\footnotesize] at (axis cs:Car,70) {66};
    \node[font=\footnotesize, xshift=2.7em] at (axis cs:Car,83) {79};

    \node[font=\footnotesize, xshift=-2.7em] at (axis cs:Person,102) {98};
    \node[font=\footnotesize] at (axis cs:Person,84) {80};
    \node[font=\footnotesize, xshift=2.7em] at (axis cs:Person,83) {79};
    \end{axis}
    \begin{axis}[
    every node near coord/.append style={font=\footnotesize, color=dark_purple, xshift=1.2em},
    ybar stacked,
    height=\barchartheight,
    width=\textwidth,
    ymin=0,
    ymax=\barchartmaxyval,
    symbolic x coords={All, Traffic Light, Stop Sign, Car, Person},
    yticklabels={,,},
    xtick={data},
    x tick label style={yshift=-\ticklabelpadding},
    bar width=\barchartwidth,
    ]
    \addplot[fill=dark_purple, nodes near coords] coordinates {
        (All, 26) (Traffic Light, 5) (Stop Sign, 18) (Car, 26) (Person, 26)
    };
    \addplot[fill=mid_purple] coordinates {
        (All, 18) (Traffic Light, 9) (Stop Sign, 9) (Car, 26) (Person, 26)
    };
    \addplot[fill=light_purple] coordinates {
        (All, 23) (Traffic Light, 27) (Stop Sign, 23) (Car, 14) (Person, 28)
    };
    \end{axis}
    \begin{axis}[
    every node near coord/.append style={font=\footnotesize, color=dark_purple, xshift=3.85em},
    ybar stacked,
    height=\barchartheight,
    width=\textwidth,
    ymin=0,
    ymax=\barchartmaxyval,
    symbolic x coords={All, Traffic Light, Stop Sign, Car, Person},
    yticklabels={,,},
    xtick={data},
    x tick label style={yshift=-\ticklabelpadding},
    bar width=\barchartwidth,
    bar shift=\barchartwidth+\barspacing,
    ]
    \addplot[fill=dark_purple, nodes near coords] coordinates {
        (All, 33) (Traffic Light, 13) (Stop Sign, 27) (Car, 38) (Person, 33)
    };
    \addplot[fill=mid_purple] coordinates {
        (All, 20) (Traffic Light, 13) (Stop Sign, 12) (Car, 23) (Person, 33)
    };
    \addplot[fill=light_purple] coordinates {
        (All, 18) (Traffic Light, 26) (Stop Sign, 17) (Car, 18) (Person, 13)
    };
    \end{axis}
\end{tikzpicture}
\caption{Attack strength of the three methods. Above the bar, the total percentage of working attacks is represented, while the percentage of strong attacks is specified on the right of each bar. }
\label{fig:attack_strength}
\end{figure*}
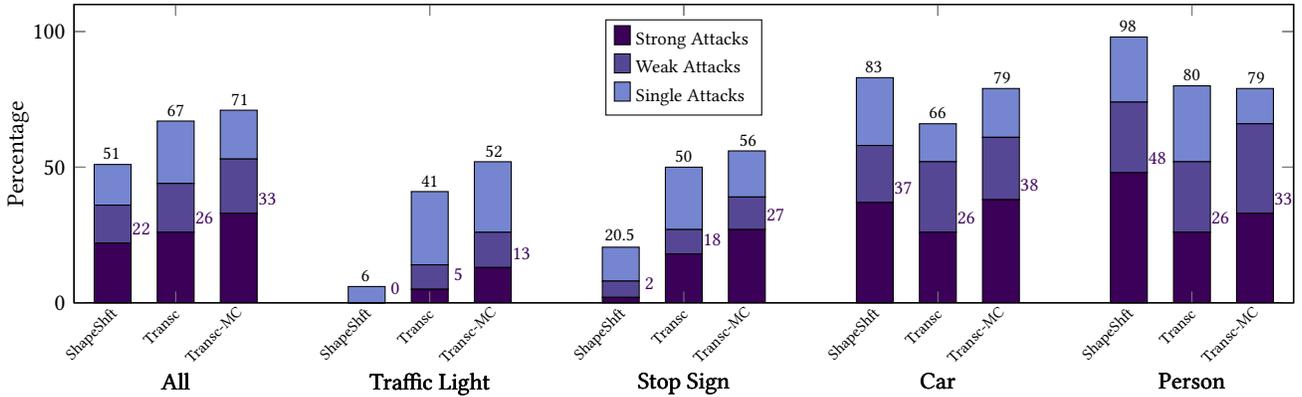

\paragraph{Transformation Configurations}
As Sava et al. \cite{sava2022assessing} motivate in their work, the robustness of an attack is dependent on the combination of transformations used for training the attack. 
However, the authors did not find a clear correlation between each individual combination and the robustness of the attack. 
Hence, in our experiments, we generate all possible transformation configurations and include all generated patches in the experimental evaluations. 

We define the following sets: 
\begin{align*}
\mathcal{S}_{\text{filters}} &= \mathcal{P}(\{\text{bright., contrast, motion blr., Gauss. noise, hue}\}) \\
\mathcal{S}_{\text{affine}} &= \mathcal{P}(\{\text{rotate, shear}\}) \\
\mathcal{S}_{\text{objects}} &= \mathcal{P}(\{\text{Barrel, Sign, Billboard}\}) - \varnothing
\end{align*}
where $\mathcal{P}(\cdot)$ is the power set. 

For ShapeShifter, the configurations set is defined as follows: 
\begin{align*}
    C_{\text{ShapeShft}} &= \{ \{\text{resize, translation}\} \cup \alpha \cup \beta \ | \\
    &\phantom{=} \ \alpha \in \mathcal{S}_{\text{filters}} \wedge \beta \in \mathcal{S}_{\text{affine}} \}
\end{align*}
and its cardinality is 128. We generate one PAE for each configuration, which results in a total of 128 adversarial examples for each target class.  

Transcender-MC and Transcender use the same transformation configurations. The set of combinations is defined as follows: 
\begin{align*}
    C_{\text{Trns-MC}} &= \{ \{\text{rotations, translations, cam. dist.}\} \cup \alpha \cup \beta \ | \\
    &\phantom{=} \ \alpha \in \mathcal{S}_{\text{filters}} \wedge \beta \in \mathcal{S}_{\text{objects}} \}
\end{align*}
which has a cardinality of 224. 
Thus, we generate 224 attacks using Transcender-MC and 224 for Transcender, for each target class.

\paragraph{Training Parameters}
For the three methods, we choose the same hyperparameters: $10^{-5}$ for the TV loss and $10^{-6}$ for the NPS. 
The patches trained with ShapeShifter undergo 15 epochs of training, Transcender patches are trained for 20 epochs and Transcender-MC attacks for 13 epochs. 
The parameters were chosen based on empirical evaluations. 
All methods generate attacks which have $608 \times 608$ pixels.  


\paragraph{The Lab Environment} 
To simulate a real-world multi-camera setup, we display the adversarial examples on a screen and place three cameras in a line in front of the attack. 
The cameras are distanced at \SI{70}{\cm} from each other to reproduce the setup presented in Subsection \ref{subsection:threat_model}.
The screen is placed in front of the central camera in three different positions: at 100, 150 and \SI{200}{\cm}. 
Since we already have three cameras with horizontal displacement, we do not additionally move the screen left or right. 
In each position, we show on the display the adversarial patches trained with the three methods.

\paragraph{Hardware}
All patches are trained on NVIDIA A100 GPUs. 
For all experiments, we used the following components: three  Logitech WebCam BRIO 4K Ultra HD for taking pictures of the patches and a Philips 24M1N3200VA 23,8 Zoll Full-HD Gaming Monitor for displaying the attacks.

\subsection{Experimental Results} 

\subsubsection{Comparative Assessment of General Attack Success Ratio of Adversarial Patches} 
This experiment investigates the effectiveness of the methods in a multi-camera setup by showcasing which percentage of the generated attacks are working or failed. 
This evaluation highlights how reliable each method is in producing robust attacks in the described scenario.
A good method generates a high percentage of strong attacks, that are able to fool all detectors, relative to the total number of working attacks. 
At the same time, an optimal method should be consistent in optimizing effective attacks, by generating a low rate of failed PAEs. 
Fig. \ref{fig:attack_strength} captures the results for this experiment. 
Each attack in each of the three positions is counted individually, such that, if an attack is strong in all three positions, we count three strong attacks. 

For all target classes, Transcender-MC has the highest rate of strong attacks, generating 11\% more patches that fooled all three cameras compared to ShapeShifter.
At the same time, our method reaches a total of 71\% working attacks, compared to ShapeShifter which stands at 51\%,
showing that Transcender-MC is the most effective in the multi-camera setup. 
While ShapeShifter fails in generating strong attacks for the classes ``traffic light'' and ``stop sign'', Transcender-MC and Transcender have a notably better performance for these two attack scenarios.  
However, PAEs optimized to be detected as the target class ``car'' yield comparable results for Transcender-MC and ShapeShifter, while the adversarial patches belonging to the class ``person'' are more effective when trained with ShapeShifter. 
The overall higher scores of the target classes ``car'' and ``person'', compared to the two other target classes, can be correlated with the class imbalance of the dataset COCO2017 by linking the under-representation with a decrease in robustness of the generated patches. 
We speculate that this behavior can be explained by the fact that the object detection model was able to define clear decision boundaries for the over-represented classes. 
Thus, because of the white-box approach, the optimization of the attacks belonging to these classes generates robust features, that are with high confidence belonging to the desired class.

\subsubsection{Comparative Assessment of General Attack Quality of Adversarial Patches} 

\begin{table}[h!]
\centering
\begin{tabular}{l|ccc}
\Xhline{2\arrayrulewidth}
\multicolumn{4}{c}{\textbf{Average Robustness Score of All Attacks}} \\
\hline \hline
\textbf{Target Cls.} & \textbf{ShapeShft.} & \textbf{Trns.} & \textbf{Trns.-MC} \\ 
\hline \hline 
All & 2.08 $\pm$ 2.48 & 2.51 $\pm$ 2.25 & \textbf{2.99 $\pm$ 2.25} \\
Traffic Light & 0.06 $\pm$ 0.17 & 0.85 $\pm$ 0.90 & \textbf{1.42 $\pm$ 1.39} \\ 
Stop Sign & 0.42 $\pm$ 0.98 & 1.82 $\pm$ 2.09 & \textbf{2.54 $\pm$ 2.22} \\ 
Car & 3.55 $\pm$ 2.58 & 2.80 $\pm$ 1.77 & \textbf{3.56 $\pm$ 2.01} \\
Person & 4.30 $\pm$ 1.77 & \textbf{4.59 $\pm$ 2.09} & 4.42 $\pm$ 2.08 \\ 
\Xhline{2\arrayrulewidth}
\multicolumn{4}{c}{\textbf{Average Robustness Score of 20\% Best Attacks}} \\
\hline \hline
\textbf{Target Cls.} & \textbf{ShapeShft.} & \textbf{Trns.} & \textbf{Trns.-MC} \\ 
\hline \hline 
All & 4.07 $\pm$ 3.11 & 5.11 $\pm$ 1.78 & \textbf{5.80 $\pm$ 1.40} \\
Traffic Light & 0.33 $\pm$ 0.24 & 2.46 $\pm$ 0.57 & \textbf{3.78 $\pm$ 0.67} \\ 
Stop Sign & 1.91 $\pm$ 1.48 & 5.42 $\pm$ 0.64 & \textbf{5.85 $\pm$ 0.79} \\ 
Car & \textbf{7.34 $\pm$ 0.50} & 5.46 $\pm$ 0.81 & 6.39 $\pm$ 0.71 \\
Person & 6.69 $\pm$ 0.43 & 7.13 $\pm$ 0.27 & \textbf{7.16 $\pm$ 0.19} \\ 
\Xhline{2\arrayrulewidth}
\end{tabular}
\caption{Average robustness score of all attacks and of best 20\% with standard deviation for ShapeShifter (ShapeShft.), Transcender (Trns.) and Transcender-MC (Trns.-MC).}
\label{table:avg_robustness_score}
\end{table}

The second experiment aims at highlighting the quality of the generated attacks. 
This can be represented by the detection score metric, which denotes the confidence of the detector that the attack belongs to the target class. 
Table \ref{table:avg_robustness_score} presents the results for this experiment, based on the average robustness score of the generated attacks. 
This score also accounts for the number of valid detections, thus attributing higher values to strong attacks. 

The first part of the table showcases the average detection scores for all attacks, underlining which method is consistent in generating patches with high robustness score.
In a real-life scenario, the attacker would try to find the patch with the highest objectness score. 
Thus, we filtered the top 20\% best attacks by robustness score for each target class.
Then, we averaged the results, obtaining the evaluations depicted in the second part of the table.

The PAEs optimized using Transcender-MC yield the highest overall scores in both scenarios (i.e. all attacks and the top 20\%). 
Mirroring the trend observed in the first experiment, the patches generated with the target class ``traffic light'' and ``stop sign'' yield lower scores than the other two groups. However, Transcender-MC and Transcender attacks reach significantly higher scores for these two target classes.
The scores for the two other target classes are comparable for Transcender-MC and ShapeShifter. 
The standard deviation of the category ``all'' is smaller for Transcender-MC and Transcender compared to ShapeShifter, this observation being accentuated for the average score of the best 20\% attacks. 
This highlights that Transcender-MC and Transcender generate patches that have similar robustness regardless of the target class.
In contrast, ShapeShifter generates effective patches only for some classes, which underlines the class-dependent reliability of the attack.


\definecolor{mid_purple}{HTML}{564795}
\definecolor{portocaliu}{HTML}{F58024}
\definecolor{verde}{HTML}{589C48}
\newcommand\linecharheight{15em}
\newcommand\linechartmaxyval{7}
\newcommand\colorss{mid_purple}
\newcommand\colortrans{portocaliu}
\newcommand\colortransmc{verde}
\begin{figure*}
    \centering
\begin{tikzpicture}
    \begin{axis}[
    height=\linecharheight,
    width=\textwidth,
    xmin=0,
    xmax=11,
    ymin=0,
    ymax=\linechartmaxyval,
    xtick={0, 1, 2, 3, 4, 5, 6, 7, 8, 9, 10, 11},
    xticklabels={Pos. 1, Pos. 2, Pos. 3, Pos. 1, Pos. 2, Pos. 3, Pos. 1, Pos. 2, Pos. 3, Pos. 1, Pos. 2, Pos. 3},
    ylabel=Robustness Score,
    y tick label style={xshift=-10pt},
    clip=false,
    legend style={cells={anchor=west}, legend pos=north west, font=\footnotesize},
    ]
    \addlegendentry{ShapeShifter}
    \addplot[color=\colorss, mark=square*, mark options={solid}, mark size=2.5pt,dashdotdotted, thick] coordinates {
        (0, 0) (1, 0.004) (2, 0.19)
    };
    \addlegendentry{Transcender}
    \addplot[color=\colortrans, mark=diamond*, mark options={solid}, mark size=3.5pt,dashed, thick] coordinates {
        (0, 0.004) (1, 0.47) (2, 2.07)
    };
    \addlegendentry{Transcender-MC}
    \addplot[color=\colortransmc, mark=pentagon*, mark options={solid}, mark size=3pt, thick] coordinates {
        (0, 0.009) (1, 2) (2, 2.25)
    };
    \addplot[color=\colorss, mark=square*, mark options={solid},mark size=2.5pt,thick,dashdotdotted] coordinates {
        (3, 0.17) (4, 0.5) (5, 0.58)
    };
    \addplot[color=\colortrans, mark=diamond*, mark options={solid}, mark size=3.5pt,thick,dashed] coordinates {
        (3, 0.2) (4, 2.53) (5, 2.72)
    };
    \addplot[color=\colortransmc, mark=pentagon*, mark options={solid}, mark size=3pt, thick] coordinates {
        (3, 0.29) (4, 3.69) (5, 3.66)
    };
    \addplot[color=\colorss, mark=square*, mark options={solid},mark size=2.5pt,thick,dashdotdotted] coordinates {
        (6, 2.71) (7, 4.39) (8, 3.56)
    };
    \addplot[color=\colortrans, mark=diamond*, mark options={solid}, mark size=3.5pt,thick,dashed] coordinates {
        (6, 0.77) (7, 4.35) (8, 3.29)
    };
    \addplot[color=\colortransmc, mark=pentagon*, mark options={solid}, mark size=3pt, thick] coordinates {
        (6, 1.25) (7, 5.55) (8, 3.9)
    };
    \addplot[color=\colorss, mark=square*, mark options={solid},mark size=2.5pt,thick,dashdotdotted] coordinates {
        (9, 3.24) (10, 3.9) (11, 5.75)
    };
    \addplot[color=\colortrans, mark=diamond*, mark options={solid}, mark size=3.5pt,thick,dashed] coordinates {
        (9, 3.16) (10, 4.43) (11, 6.18)
    };
    \addplot[color=\colortransmc, mark=pentagon*, mark options={solid}, mark size=3pt, thick] coordinates {
        (9, 3.58) (10, 4.98) (11, 4.69)
    };

    \node at (axis cs:1,-1.37) {\textbf{Traffic Light}};
    \node at (axis cs:4,-1.37) {\textbf{Stop Sign}};
    \node at (axis cs:7,-1.37) {\textbf{Car}};
    \node at (axis cs:10,-1.37) {\textbf{Person}};

    \end{axis}
\end{tikzpicture}
\caption{Average robustness score of the patches, represented for each of the camera distances. In position one, the camera was at 150 cm from the display, in position two, at 100 cm and in the third position, at 50 cm.}
\label{figure:avg_robustness_score_positions}
\end{figure*}
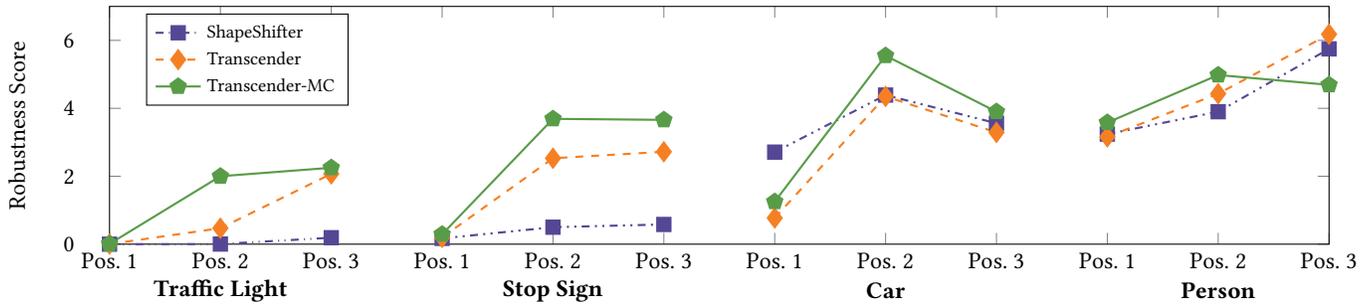

\subsubsection{Comparative Assessment of General Attack Quality Based on Position}

The effectiveness of the patch is also strongly influenced by the distance between the cameras and the attack. 
This evaluation highlights the correlations between the position and the effectiveness of the patch by computing the average robustness score in each of the three positions. 
Fig. \ref{figure:avg_robustness_score_positions} presents the results.

Transcender-MC has the highest score among the three methods in all positions, with only two exceptions from the total of 12, showcasing that it is the most reliable in optimizing PAEs that are effective regardless of the position. 
At the same time, the influence of the target class is visible on the average robustness score in each position. 
While the patches optimized with target class ``traffic light'' yield the highest robustness scores in the closest position to the camera, the adversarial examples of the class ``car'' reach the best scores in the second position, for all methods. 
However, it is clear that, for all target classes and methods, the effectiveness of the patches faces a drop in the farthest position. 
This is presumably based on the camera resolution, which is not able to capture relevant features at distance. 

\subsubsection{Comparative Assessment of Adversarial Patches in Difficult Conditions} \label{subsection: difficult scenarios}
\begin{figure}
    \centering
       \subfloat[]{\includegraphics[height=0.08\textheight]{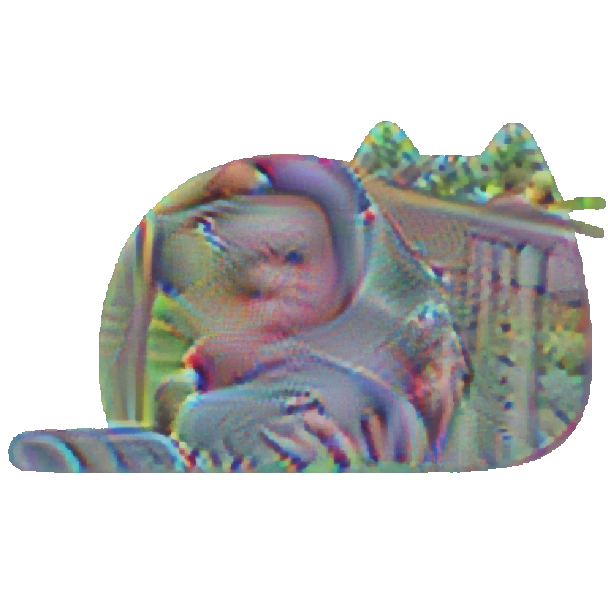}}
    \hspace{8px}
    \subfloat[]{\includegraphics[height=0.08\textheight]{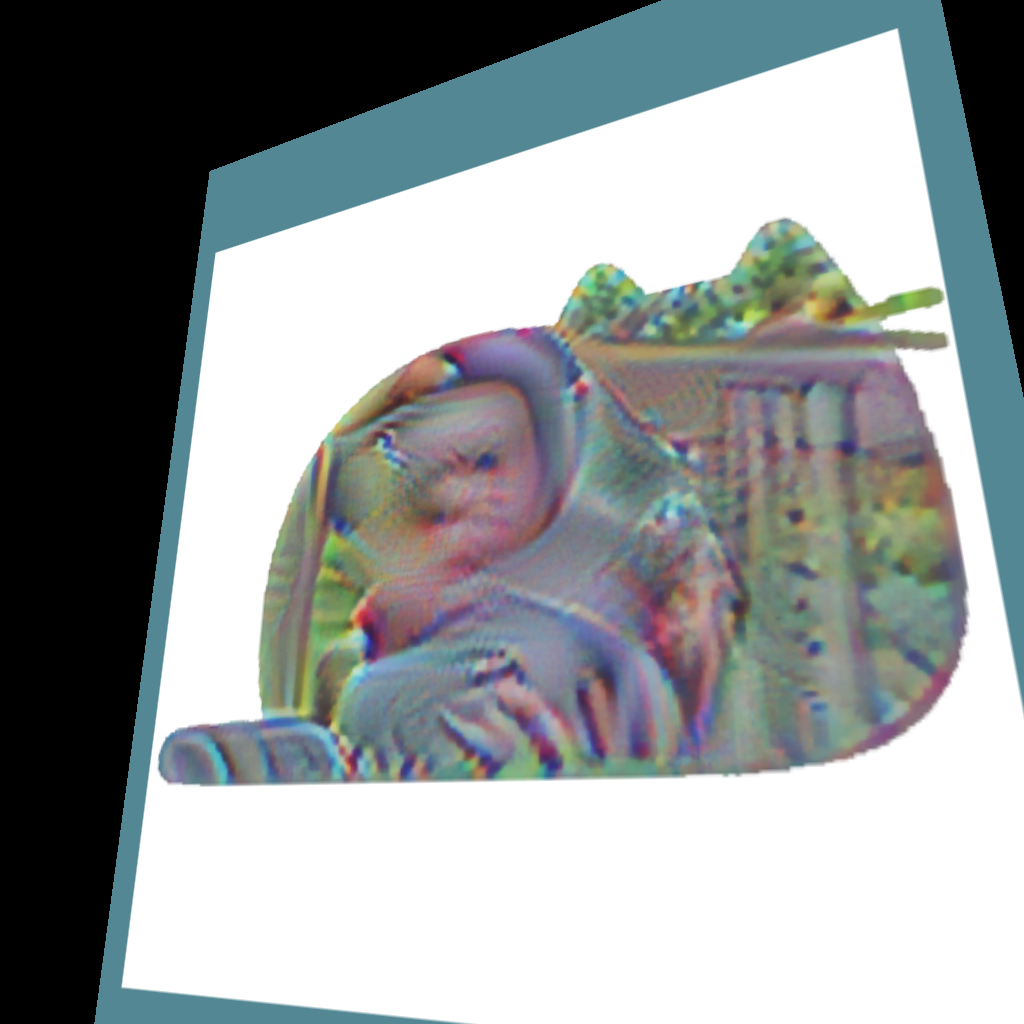}}
    \hspace{8px}
    \subfloat[]{\includegraphics[height=0.08\textheight]{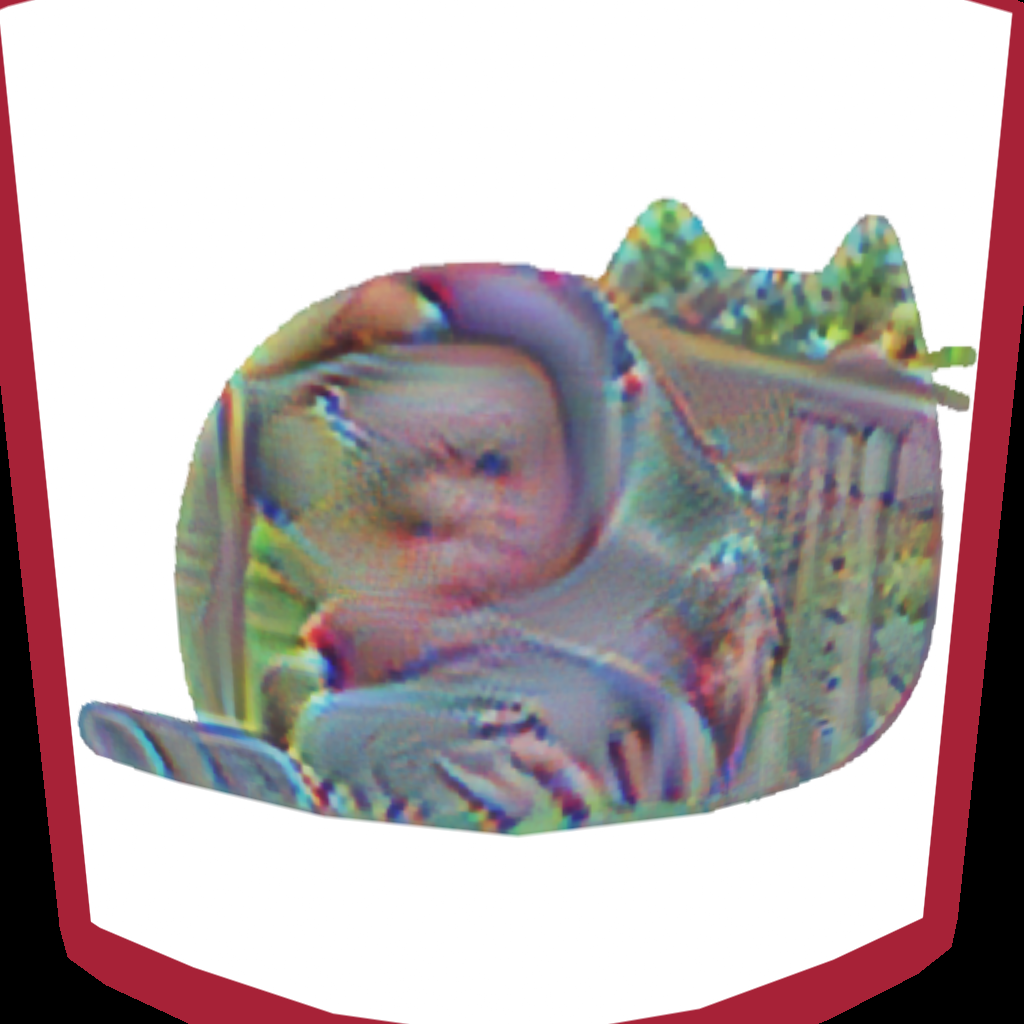}}
    \hspace{8px}
    \subfloat[]{\includegraphics[height=0.08\textheight]{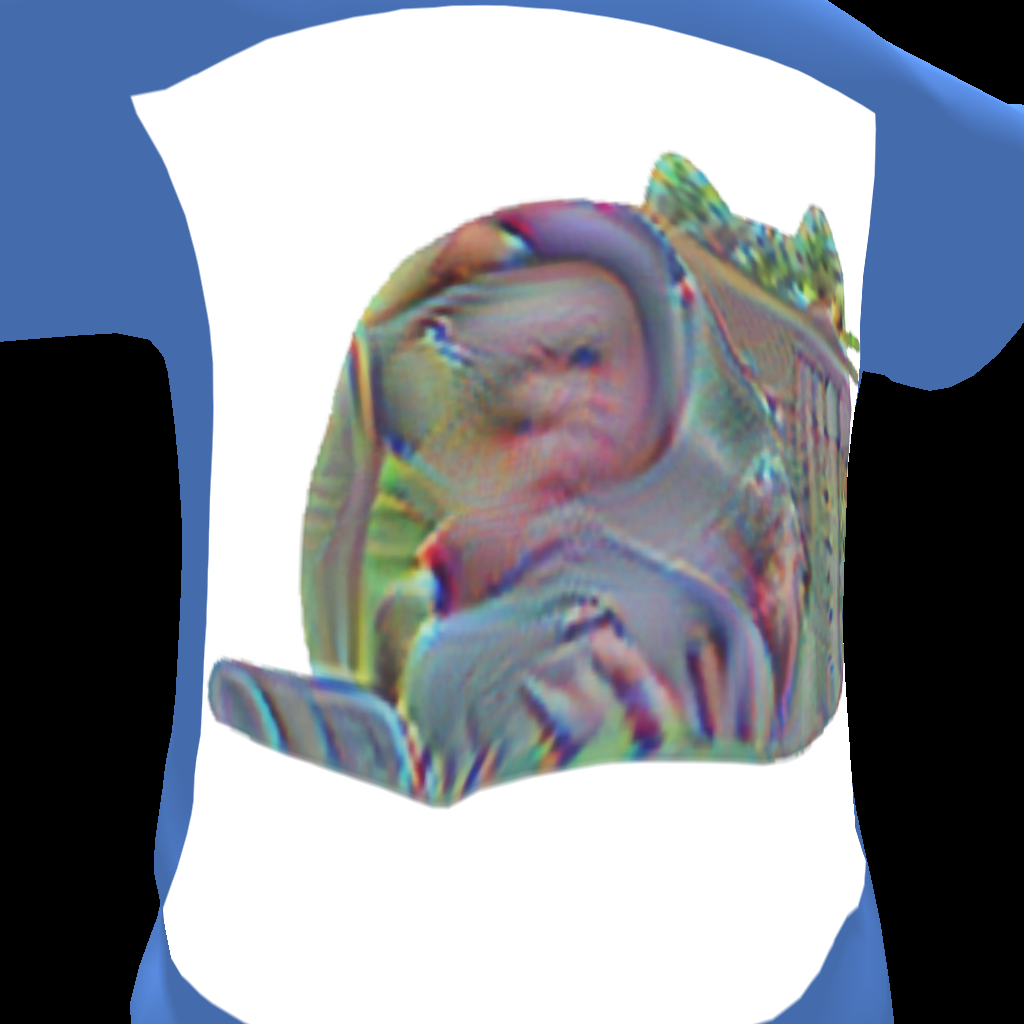}}
    \caption{Adversarial attack with target class ``person'' (a), and the attack applied on the billboard (b), on the traffic sign (c) and on the t-shirt (d). The objects are rotated at small random angles. To increase the stealthiness of the attack, we apply a cat emoji mask during training.}
    \label{fig:3d_rendering}
\end{figure}

In a physical setup, the planar attacks need to be placed on a support object from the physical environment, such as cardboard sheets, a traffic sign, or on pieces of clothing. 
Since each object has a different surface, the attack might undergo deformations.
Moreover, the cameras are not front-facing the attack, but are rotated at random angles. 
Hence, in this experiment we investigate the robustness of the patches in difficult conditions. 
We use a 3D renderer to apply the adversarial patches on different 3D meshes. 
For this, a 3D object is first selected.
Next, for each generated PAE, we proceed as follows: (1) we select a random texture color, (2) we apply the attack on the object in a predefined position, (3) we rotate the camera around the object to simulate the rotation of the object itself, and (4) we render the images. 
We perform this procedure for three different support objects: a billboard and a sign, used at training for Transcender-MC, and a new 3D object, a t-shirt. 
Fig. \ref{fig:3d_rendering} highlights the results of this process, showing a generated attack printed on all three objects.  
All rendered images are displayed on the screen in each of the three positions, and the images are captured with the three cameras as presented above.

Table \ref{table:attack_success_ratio_difficult} displays the percentage of strong attacks for each of the three objects. 

Transcender-MC generates the highest percentage of strong attacks for all target classes, for the support objects ``billboard'' and ``sign''. 
The two objects were also used as training 3D meshes for this method. 
This showcases that Transcender-MC is the most reliable method in creating robust attacks, as long as the attacker has access to a 3D virtual copy of the object used as support in the physical world, in order to include it in the 3D mesh set during training. 

On the other hand, patches trained with ShapeShifter face a major cutoff in robustness in the ``billboard'' attack scenario. 
Compared to the first experiment, the percentage of strong patches trained with ShapeShifter belonging to the target class ``car'' was reduced by 30\% (37\% strong attacks in first experiment, 7\% in current experiment). Similarly, the PAEs trained with the class ``person'' faced a steep drop in robustness, yielding 45\% less strong attacks (48\% in first experiment, 3\% in current exoeriment). 
Since the billboard is planar, it does not deform the patch, meaning that the rotations alone have a negative influence on the robustness. 
This is an empirical result which stands in agreement with prior work \cite{Lennon_2021_ICCV, tarchoun_2, tarchoun_1}. 
However, Transcender-MC makes the patches more robust to out-of-plane 3D rotations. 

For the support object ``t-shirt'', Transcender-MC shows the highest percentages for the classes ``stop sign'' and ``person'', however the patches generated using ShapeShifter are more robust for the two other classes, highlighting that Transcender-MC patches also experience a drop in performance when applied on new objects. 

While in the previous experiments the patches trained with ShapeShifter using the target class ``person'' have the highest rate of successfully attacking all three detectors, this advantage is lost in the current attack scenario. 
In this experiment, only small percentages of the patches were strong for the aforementioned target class (3\% for ``billboard'', 20\% for ``sign'' and 3.9\% for ``t-shirt''). 
This is an empirical evidence that, in complex scenarios, ShapeShifter can lose its benefit of optimizing effective attacks for some target classes. 
On the other hand, Transcender-MC excelled in all three cases for the target class ``person'', generating a high percentage of strong attacks that are robust to 3D angles and deformations.

\begin{table}[h!]
\centering
\begin{tabular}{l | ccc}
\Xhline{2\arrayrulewidth}
\multicolumn{4}{c}{\textbf{Percentage of Strong Attacks: Billboard}} \\
\hline \hline
\textbf{Target Cls.} & \textbf{ShapeShft.} & \textbf{Trns.} & \textbf{Trns-MC.} \\ 
\hline \hline 
Traffic Light & 0\% & 0\% & \textbf{2\%} \\ 
Stop Sign & 0\% & 2\% & \textbf{6\%} \\ 
Car & \textbf{7\%} & 2\% & \textbf{7\%} \\
Person & 3\% & 17\% & \textbf{22\% }\\ 
\Xhline{2\arrayrulewidth}
\multicolumn{4}{c}{\textbf{Percentage of Strong Attacks: Sign}} \\
\hline \hline
\textbf{Target Cls.} & \textbf{ShapeShft.} & \textbf{Trns.} & \textbf{Trns-MC.} \\ 
\hline \hline 
Traffic Light & 0.5\% & 0.3\% & \textbf{4\%} \\ 
Stop Sign & 0\% & 4\% & \textbf{8\%} \\ 
Car & 12\% & 9\% & \textbf{14\%} \\
Person & 20\% & 31\% & \textbf{40\%} \\ 
\Xhline{2\arrayrulewidth}
\multicolumn{4}{c}{\textbf{Percentage of Strong Attacks: T-Shirt}} \\
\hline \hline
\textbf{Target Cls.} & \textbf{ShapeShft.} & \textbf{Trns.} & \textbf{Trns.-MC} \\ 
\hline \hline 
Traffic Light & \textbf{2.86\%} & 1.63\% & 2.5\% \\ 
Stop Sign & 0\% & 0.6\% & \textbf{0.9\%} \\ 
Car & \textbf{0.7\%} & 0\% & 0.1\% \\
Person & 3.9\% & 11\% & \textbf{15\%} \\ 
\Xhline{2\arrayrulewidth}
\end{tabular}
\caption{Percentage of strong detections by support object.}
\label{table:attack_success_ratio_difficult}
\end{table}

\subsection{Ablation Studies}
\subsubsection{Assessment of Training Data Augmentation in Transcender-MC} 
Transcender-MC is designed to perform well in a multi-camera setup. 
Hence, we evaluate this method against Transcender, which is the data-augmentation-free variant of the method. 
This comparison highlights the impact of the augmentation technique on the robustness of patches in a multi-camera setup.

Regarding the attack strength, according to Fig. \ref{fig:attack_strength}, Transcender-MC generated 46\% strong attacks with respect to all working attacks (33\% strong attacks and 71\% working attacks relative to all attacks), and Transcender, 38\% (22\% strong attacks and 51\% working attacks relative to all attacks). 
This underlines that, not only did Transcender-MC generate more working attacks, but also a higher percentage of them are strong, showing the positive influence of the training data augmentation technique on the multi-camera robustness of the patches. 

In terms of attack quality, presented in Table \ref{table:avg_robustness_score}, Transcender-MC has a higher overall average robustness score for the category ``all'' and for the best 20\% of the patches. 
Transcender-MC brings an increase of 0.48 for all attacks and 0.69 for the best 20\% in robustness score compared to Transcender, depicting the positive influence of the training data augmentation method.   

Patches trained using Transcender-MC are also more effective in complex scenarios in comparison to those trained with Transcender. 
According to Table \ref{table:attack_success_ratio_difficult}, for all target classes and support objects, Transcender-MC has higher rates of strong attacks.
Moreover, in this experiment, Transcender alone is not consistent in delivering better results compared to ShapeShifter. 
Hence, the multi-camera augmentation makes the planar patches more robust to out-of-plane rotations and deformations, which is fundamental in a real-life attack scenario.

\subsubsection{Assessment of the Impact of 3D Renderings in Transcender}
To evaluate the effect of including 3D renderings of objects as support for the patches during training, we refer to a comparative assessment of ShapeShifter and Transcender. 

With regard to strong attacks (Fig. \ref{fig:attack_strength}), Transcender has a higher rate, standing at 26\%, compared to ShapeShifter, which only yields 22\% strong attacks. 
However, by inspecting the results for the individual target classes, we observe that this difference is accumulated from the two target classes ``traffic light'' and ``stop sign'', that have considerably higher rates for Transcender. 
On the other hand, ShapeShifter yields higher strong attack rates for the two other target classes. 
This highlights that the use of 3D meshes for training has a beneficial impact on the robustness of patches trained with target classes that generate non-effective PAEs with ShapeShifter.   

On the other hand, according to Table \ref{table:avg_robustness_score}, even though ShapeShifter generated a higher percentage of strong attacks for the class ``person'' compared to Transcender (48\% for ShapeShifter and 26\% for Transcender), the patches generated with Transcender for this class obtained a higher robustness score (4.30 for ShapeShifter, 4.59 for Transcender for all attacks and 6.69 for ShapeShifter, 7.13 for Transcender for the best 20\%). 
This implies that Transcender plays a positive role in increasing the objectness score of the attacks, even if the optimized patches are not strong. 

\subsubsection{Evaluation of the Impact of Training 3D Object Pools}
The training configuration of Transcender-MC and Transcender includes the choice of a 3D object pool that the objects are applied on.
At each iteration step, one of the objects from the pool is selected. 
This evaluation has the purpose of revealing the impact of the object pool on the robustness of the adversarial patches. 
For this, we use the detection results from the difficult scenario that is described in Subsection \ref{subsection: difficult scenarios}: all the trained PAEs are pasted on different support objects and random camera rotations are applied. 
In this ablation study, we differentiate between the training object pool, and the support object --- the 3D mesh on which we apply the trained patches at test time.
\begin{table}[h]
\centering
\begin{tabular}{l||c|c||c|c||c|c}
\Xhline{2\arrayrulewidth}
\multicolumn{7}{c}{\textbf{Support Object Distribution for Transcender-MC}} \\
\hline \hline
\textbf{Support Object} &\multicolumn{2}{c}{\textbf{Billboard}} & \multicolumn{2}{c}{\textbf{Sign}} & \multicolumn{2}{c}{\textbf{T-shirt}} \\ 
\hline \hline 
\textbf{Training Obj.} & \textbf{Neg.} & \textbf{Pos.} & \textbf{Neg.} & \textbf{Pos.} & \textbf{Neg.} & \textbf{Pos.} \\
\hline
Billboard &\textbf{ 48\%} & 9\% & 45\% & \textbf{22\%} & 76\% & 4\%  \\ 
Sign & 57\% & 6\% & 46\% & 15\% & 77\% & 3\% \\ 
Barrel & 50\% & 7\% & 53\% & 13\% & 78\% & 4\%  \\ 
Billboard, Sign & \textbf{48\%} & \textbf{13\%}& \textbf{44\%} & 16\% & \textbf{69\%} & 5\% \\ 
Billboard, Barrel & \textbf{48\%} & 11\% & 49\% & 16\% & \textbf{69\%} &\textbf{ 6\% }\\ 
Sign, Barrel & 54\% & 9\% & 45\% & 16\% & 72\% & 3\% \\ 
Billb., Sign, Barrel & \textbf{48\%} & 10\% & 45\% & 15\% & 73\% & 5\% \\
\Xhline{2\arrayrulewidth}
\end{tabular}
\caption{Training objects distribution in failed attacks (``Neg.'') and strong attacks (``Pos.'') for Transcender-MC, for each support object.}
\label{table:object_distribution_transcender-mc}
\end{table}

Table \ref{table:object_distribution_transcender-mc} shows the impact of each training object pool on the effectiveness of attacks based on the support object. 
For each support object, it depicts the percentage of strong and failed attacks that are trained using each object pool.
We consider a group of training meshes to be beneficial for the robustness, if a high percentage of attacks optimized with it are strong (denoted with ``Pos.'' in Table \ref{table:object_distribution_transcender-mc}) and a low percentage of attacks are failed (denoted with ``Neg.''), relative to other object pools.
According to the table, the groups of training meshes, that are most beneficial for the robustness, are ``billboard'', ``billboard, sign'' and ``billboard, barrel''. 
For the support object ``billboard'', the training set ``billboard, sign'' was present in 48\% of the failed patches and 13\% of the positive patches.
In contrast, the ``sign'' as training mesh generated the worst results, yielding a high number of failed patches (57\%) and a low number of positive ones (6\%). 
For the support object ``sign'', the ``billboard, sign'' group had the lowest appearance frequency in failed patches (44\%), while the ``billboard'' as training object had the highest rate of appearance in positive patches (22\%). 
This shows that, besides using an object with a flat surface such as a billboard, including meshes with non-planar shapes helps at increasing the robustness in a multi-camera setup. 
A substantial positive difference in results is visible for the ``t-shirt'' support object when the training mesh pools ``billboard, sign'' and ``billboard, barrel'' are used. 
The two groups have the lowest rate of appearance of 69\% in failed patches.
The ''billboard, barrel'' combination was also beneficial for training robust patches for the support object ``t-shirt'', having a positive appearance of 6\%, while other groups have lower rates of only 3-5\%. 
This finding suggests that, especially when printing the attack on new support objects, including meshes with various shapes in the training pool has an advantageous effect for the robustness. 

\subsection{Future Work} 
In this work, we presented a preliminary study on physical adversarial example against a multi-camera system. Our physical camera setup was comprised of three video devices, each camera forwarding the video input to a 2D object detector.
A task for future work is analyzing the attacks presented here on a different physical setup, that uses a varied number of cameras or employs a 3D object detector.
Moreover, the results and observations presented in our work can be compared with the results obtained with different setups. This would help in determining which setup offers the best protection against adversarial patches.
We also introduced Transcender-MC, which makes use of 3D renderings of meshes for increasing the robustness of the attack.
While we conducted an initial study on the influence of 3D mesh pools, additional investigations are needed to discover which objects could make the generated patches transferable to random surfaces in the physical world. 

\section{Conclusion}
In this paper, we evaluated the robustness of a multi-camera setup against adversarial patches. 
We propose a new method for generating physical attacks, called Transcender-MC, which leverages differentiable rendering to improve the robustness of physical planar attacks.
Our proposed method also includes a training data augmentation technique that makes patches more effective in a setup with multiple cameras. 
We perform a comparative evaluation of Transcender-MC and another SOTA method against a setup with three parallel cameras that resembles the configuration of video devices on the front of an autonomous vehicle. 

We empirically show that Transcender-MC is better at generating robust attacks in a multi-camera setting in our evaluations. 
The results suggest that our proposed method is capable of optimizing a higher rate of attacks that fool all cameras from our setup, and the generated patches also yield higher average detection scores. 

We discover that patches trained with Transcender-MC are significantly more robust for target classes that ShapeShifter encounters difficulties in optimizing. 
Additionally, under certain circumstances, Transcender-MC generates PAEs that are robust in difficult conditions, under out-of-plane camera angles and deformations.

Overall, our study highlights that, while a multi-camera setup offers some protection against adversarial examples, such a system can still be vulnerable if the training pipeline is adapted to optimize for a multi-camera, three-dimensional environment. 
Through this, it offers significant insights regarding the potential threat represented by 2D adversarial examples and encourages further development of defense mechanisms for platforms that employ multiple video devices.

\bibliographystyle{ACM-Reference-Format}
\bibliography{main}

\appendix
\section{Appendix}

\subsection{Training Parameters for ShapeShifter} 
ShapeShifter uses two types of transformations during training: image distortions and affine transformations. Table \ref{table:shapeshifter_parameters} presents the parameter choice for affine transformations that we used to optimize this method. The image filters that we applied coincide with those of Transcender-MC, and the parameter values are presented in Table \ref{table:all_parameters}.

\begin{table}[h]
\centering
\begin{Tabular}[1.2]{l|lll}
\Xhline{2\arrayrulewidth}
\textbf{Aff. Trans.} & Mechanism & Param. & Range \\
\hline \hline 
 Resize & $A_{\alpha, \beta}X$ & $\alpha$: scal.& $[\frac{40}{X_{h}}, \frac{180}{X_{h}}]$ \\
& & $\beta$: scal. & $[\frac{5}{X_{w}}, \frac{60}{X_{w}}]$ \\
Translate & $X + T_{\alpha, \beta}$ & $\alpha$: scal. & $[-\delta_{h}, \delta_{h}]$  \\
& & $\beta$: scal. & $[-\delta_{w}, \delta_{w}]$ \\
Rotate & $R_{\theta}X$ & $\theta$: angle & $[-10^{\circ}, 10^{\circ}]$ \\
Shear & $S_{\theta}X$ & $\theta$: angle & $[0^{\circ}, 15^{\circ}]$ \\
\Xhline{2\arrayrulewidth}
\end{Tabular}
\caption{
    Parameter choice for affine transformations of ShapeShifter. $X$ is the input adversarial patch.}
\label{table:shapeshifter_parameters}
\end{table}

\end{document}